\documentclass{JUSTC}

\usepackage{algorithm,algorithmicx,algpseudocode}%
\usepackage{multirow}
\usepackage{tabularx}
\usepackage{booktabs}
\usepackage{subcaption}

\type{Manuscript type}
\title{Geometry-aware Gaussian Prior and Axial Attention for Cervical Cytology Image Classification}

\author[1]{Yating Li}

\author[2]{Cheng Ye}

\author[3,\Letter]{Nenan Lyu}

\author[2,\Letter]{Weidong Chen}

\author[1]{Zhendong Mao}

\email{lyunenan@cicams.ac.cn}

\email{chenweidong@ustc.edu.cn}

\runningtitle{Geometry-aware Cytology Classification}
\runningauthor{Yating Li \etal}

\address{School of Cyber Science and Technology, University of Science and Technology of China, Hefei 230026, China}

\address{School of Information Science and Technology, University of Science and Technology of China, Hefei 230026, China}

\address{Department of Gynecologic Oncology, National Cancer Center, National Clinical Research Center for Cancer, Cancer Hospital, Chinese Academy of Medical Sciences and Peking Union Medical College, Beijing 100730, China}

\abstract{
  Accurate cervical cytology image classification is a key component of automated cervical cancer screening, where reliable recognition of normal, precancerous, and cancer-associated cellular patterns from Pap smear images can improve screening efficiency and diagnostic consistency. However, this task remains challenging because cervical cells exhibit complex morphology, subtle intra-class variations, and strong inter-class similarities. Existing convolution-based models capture local texture well but have limited ability to model long-range relationships, whereas attention-based models provide broader context but often lack explicit structural guidance. To address these limitations, we propose a geometry-aware classification framework for cervical cancer screening-oriented cytology image analysis, incorporating semantic abstraction and structural priors learned from pre-trained vision-language features. The method uses Gaussian expert modules to generate axis-wise priors from global semantic information, capturing structural regularities such as nuclear alignment and cellular spatial organization. These priors are embedded into an axial self-attention module to modulate similarity computation along horizontal and vertical directions, improving long-range dependency modeling and structure-sensitive feature interaction. Experiments on the Mendeley liquid-based cytology and SIPaKMeD datasets show that the proposed method achieves 99.48\% accuracy on the former and 96.08\% on the latter, with balanced gains in recall, precision, and overall classification performance. Visual analysis further shows that the learned priors highlight diagnostically relevant cellular regions, demonstrating the potential of the proposed framework as a screening-oriented decision-support tool for cervical cytology.
}


\begin{document}

\maketitle 

\section{Introduction}

Cervical cancer is one of the most common gynecological malignancies worldwide, with both high incidence and mortality rates, particularly in low- and middle-income countries. According to statistics, there were over 600,000 newly diagnosed cases and more than 300,000 deaths globally in 2020 \cite{world2020global}. At present, diagnosis primarily relies on expert pathologists manually examining Pap smear slides under a microscope. This process is time-consuming, labor-intensive, and subject to inter-observer variability, which increases the risk of misdiagnosis. Consequently, the development of efficient and reliable computer-aided diagnosis (CADx) systems has become an urgent necessity. In this context, automated cervical cytology image classification has attracted significant attention as a key component of computer-assisted cervical cancer screening, since accurate recognition of abnormal cellular morphology can improve screening efficiency, reduce inter-observer variability, and provide reliable decision support for cytological assessment.
Compared with conventional computer vision tasks, screening-oriented cervical cytology classification poses unique challenges, as it requires distinguishing subtle morphological variations at the cellular and subcellular levels that are closely associated with normal, precancerous, and cancer-related cytological patterns.Previous approaches to cervical cytology classification have generally relied on deep networks to extract hierarchical representations of cellular morphology, sometimes combined with dimensionality reduction, feature aggregation, or classifier fusion to enhance discrimination \cite{7932065, basak2021cervical, dosovitskiy2020image, shao2021transmil}. While these strategies have achieved encouraging progress, they still face significant limitations. In particular, they often fail to adequately capture slice-level structural information, lack the sensitivity to fine-grained morphological variations critical for diagnosis, and remain highly dependent on large annotated datasets and handcrafted preprocessing steps. Moreover, the absence of explicit geometric priors results in weak structural awareness and limited interpretability. Collectively, these shortcomings constrain the robustness and applicability of existing cytology classification methods in real-world clinical practice.

\begin{figure}[t]
    \centering
    \includegraphics[width=\linewidth]{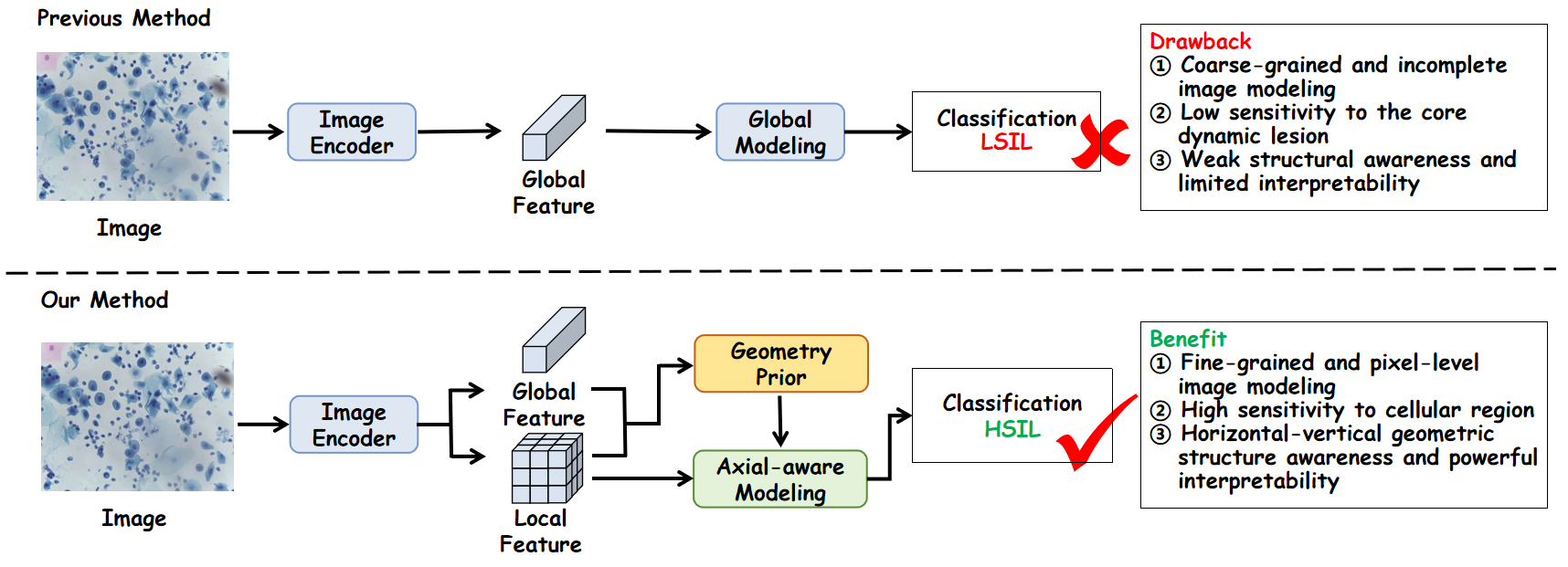}
    \caption{
        Comparison between previous cervical cytology classification pipelines and our proposed framework. 
        Previous methods rely solely on global modeling, leading to coarse-grained representations, weak structural awareness, and limited interpretability. 
        In contrast, our method incorporates geometry-aware priors and axial structural modeling, enabling fine-grained feature extraction, enhanced sensitivity to cellular regions, and improved diagnostic reliability.
    }
    \label{fig:intro_comparison}
\end{figure}

To visually illustrate these limitations and our improvements, Fig.~\ref{fig:intro_comparison} provides a brief comparison.

To address these challenges, we propose a novel cervical cancer screening-oriented cytology classification framework that embeds domain-inspired inductive biases directly into network design to enable more robust and interpretable Pap smear image analysis. Specifically, we adopt the pretrained CLIP model as a feature extractor, leveraging its strong semantic representation capability acquired through large-scale multimodal pretraining. This choice is consistent with recent multimodal vision studies showing that semantic abstraction and cross-modal reasoning can benefit medical report generation, image captioning, and structure-aware visual understanding \cite{chen2024bootstrapping, jin2024multigrained, jin2024d2net, wang2023structuredconcepts, li2024visualrelationships}. Building on this foundation, we introduce two task-specific modules that jointly capture global semantics and local structural features.

The first innovation is the Geometry-aware Gaussian Prior Module, which introduces a set of controllable Gaussian expert functions to model the spatial and directional distribution of cells. Guided by global semantic features, this module dynamically generates direction-sensitive weighting patterns, emphasizing diagnostically relevant regions. Compared with traditional attention mechanisms, it enhances the model’s sensitivity to structural patterns and improves its ability to discriminate subtle morphological variations, thus providing a stronger inductive bias for cytology classification.

The second innovation is the Gaussian-enhanced Axial Attention Mechanism. This module incorporates Gaussian priors into the axial attention framework, enforcing directional consistency and structural constraints along the horizontal and vertical dimensions. Unlike standard self-attention, it simultaneously captures long-range dependencies and retains high sensitivity to fine-grained structural differences, while significantly reducing computational complexity. This design not only improves the efficiency of processing high-resolution cytology images but also enhances the adaptability of the overall framework to complex diagnostic scenarios.

By integrating Gaussian priors with axial attention in a unified architecture, our framework effectively balances global semantic modeling and local structural representation. Experiments on multiple public cervical cytology datasets demonstrate that our method outperforms state-of-the-art baselines in accuracy, generalization, and robustness, highlighting its strong potential for clinical application. The main contributions of this paper are summarized as follows:

\begin{itemize}
\item We propose a novel cervical cancer screening-oriented cytology classification framework that integrates geometric priors with structure-aware attention, achieving a unified balance between global semantic modeling and local structural representation.
\item We design a Geometry-aware Gaussian Prior Module that adaptively captures the spatial and directional distribution of cells, enhancing structural sensitivity and clinical discriminative focus.
\item We develop a Gaussian-enhanced Axial Attention Mechanism that enforces direction-sensitive structural consistency while reducing computational complexity, thereby improving sensitivity to subtle pathological differences.
\item Extensive experiments and ablation studies on multiple public cervical cytology datasets demonstrate that our method achieves superior accuracy, generalization, and robustness compared with existing methods, underscoring its strong potential for cervical cancer screening-oriented clinical deployment.
\end{itemize}

\section{Related Work}

\subsection{Medical Image Classification}

Medical image classification is a critical component of computer-aided diagnosis, aiming to automatically extract discriminative features from complex and noisy medical images and accurately assign disease categories. With the advancement of deep learning, end-to-end trainable networks have shown strong performance across diverse clinical applications, significantly improving the efficiency and consistency of diagnosis.  

Convolutional neural networks (CNNs) have long dominated image classification since the success of AlexNet on ImageNet \cite{krizhevsky2012imagenet}. Subsequent designs, such as Network in Network (NIN) \cite{lin2014networknetwork}, GoogLeNet \cite{Szegedy_2015_CVPR}, and VGG \cite{simonyan2014very}, progressively increased the depth and width of networks to capture richer hierarchical features. The introduction of residual connections in ResNet \cite{He_2016_CVPR} addressed the vanishing gradient problem and enabled the stable training of very deep networks. In the medical domain, CNN-based models have been successfully applied to various diagnostic tasks, including retinal vessel segmentation \cite{7440871}, skin lesion classification \cite{Esteva2017}, and pneumonia detection in chest X-rays \cite{rajpurkar2017chexnetradiologistlevelpneumoniadetection}. Later advances such as DenseNet \cite{huang2017densely} and EfficientNet \cite{tan2019efficientnet} introduced dense connectivity and compound scaling strategies, further enhancing classification accuracy and parameter efficiency. Moreover, hybrid models like Fine\_Denseiganet \cite{sahu2025finedenseiganet}, which combine fine-grained feature extraction with DenseNet modules, have shown strong robustness on chest CT datasets, underscoring the effectiveness of multi-component architectures for complex medical image analysis.  

Recently, Transformer-based models have gained significant traction in vision tasks \cite{NIPS2017_3f5ee243}. The Vision Transformer (ViT) \cite{dosovitskiy2020image} partitions images into fixed-size patches and uses self-attention to capture global dependencies, achieving results competitive with CNNs on large-scale datasets. Building upon this, variants such as DeiT \cite{touvron2021training} introduced data-efficient training strategies, while DynamicViT \cite{rao2021dynamicvit} improved computational efficiency by pruning redundant tokens. In the medical domain, Transformer architectures have been explored for multimodal fusion, multi-instance learning, and large-scale whole-slide image analysis. For example, TransMed \cite{dai2021transmed} performs joint modeling of heterogeneous medical modalities, TransMIL \cite{shao2021transmil} tackles whole-slide image classification with instance-level aggregation, and Lesion-aware Transformers \cite{sun2021lesion} incorporate localized attention for subtle lesion detection. More recently, Swin Transformer \cite{liu2021swin} and its medical adaptations \cite{hatamizadeh2021swin} demonstrated the effectiveness of hierarchical Transformer backbones for 3D medical image analysis. These works collectively highlight the growing role of Transformers in modeling complex structural and semantic patterns in medical images.  

In summary, CNN-based and Transformer-based approaches have significantly advanced medical image classification. However, challenges remain in balancing global semantic modeling and fine-grained structural perception, motivating the exploration of hybrid designs and domain-inspired inductive priors.  

\subsection{Multimodal Vision and Structure-aware Representati}

Beyond conventional classification backbones, recent multimodal learning studies have emphasized that visual representations can be strengthened by semantic guidance, cross-modal interaction, and structured reasoning. In the medical domain, radiology report generation has benefited from large language model bootstrapping, multi-grained abnormality prediction, and diffusion-enhanced discriminative modeling \cite{chen2024bootstrapping, jin2024multigrained, jin2024d2net}. Although these works focus on report generation rather than cytology classification, they support the broader view that clinically meaningful visual patterns should be modeled jointly with high-level semantic abstraction.

Similar ideas have also been explored in general visual-language understanding. Image captioning methods have incorporated structured concepts, transformer-based visual relation graphs, and contour-aware concept prediction to improve the alignment between visual regions and semantic descriptions \cite{wang2023structuredconcepts, li2024visualrelationships, wang2023contour}. In natural language and multimodal reasoning tasks, type-aware multi-hop question generation, contrastive transfer pattern mining, and grammar-aware sentiment analysis further demonstrate the value of explicit structure in semantic modeling \cite{lin2024fewshot, han2023contrastive, tian2023aspect}. These studies motivate our use of pretrained semantic features as a high-level guide for learning structure-sensitive cytology representations.

For video and temporal visual understanding, prior work has examined sentence-guided actor-action segmentation, compressed-video referring object segmentation, weakly supervised text-based video segmentation, and partially relevant video retrieval \cite{chen2021cascade, chen2022multiattention, chen2023weakly, song2025retrieval}. More recent studies extend this line to emotional video captioning, subjective video captioning, affective explanation captioning, multimodal empathetic response generation, retrieval-enhanced emotional captioning, and efficient audio-visual captioning \cite{ye2024dualpath, ye2025videosummarization, ye2025multiround, chen2026subjective, zhang2026stimuli, song2026bridging, wang2026multiagent, chen2026facenet, chen2026emotionattributed, meng2026audiovisual}. These works are complementary to ours because they highlight the importance of long-range dependency modeling and fine-grained visual evidence, both of which are also central to screening-oriented cytology image analysis.

In addition, structure-aware multimodal representation learning has recently expanded to anomaly detection, creative graphic design generation, composed image retrieval, micro-action recognition, audio-visual question answering, stance detection, visual emotion dataset construction, cross-view image matching, key relation detection, and live video commenting \cite{huang2025graphmoe, zhang2025creatidesign, hong2025creatiposter, chen2026creatiparser, li2025pseudo, wang2025microaction, qin2025tokenpruning, zhou2025stance, guo2025emoverse, liu2025street, zhao2023difference, fu2024sentimentvae}. Collectively, these studies suggest that task-specific priors and semantic routing mechanisms can improve representation quality across diverse visual scenarios. Our method adapts this insight to cervical cytology by using Gaussian experts to inject direction-sensitive geometric priors into axial attention.

\subsection{Geometric Priors}

Geometric priors provide explicit or implicit structural constraints that guide deep networks to better capture local details, global dependencies, and multi-scale variations. By embedding such priors, models gain stronger inductive biases, improving both robustness and interpretability.  

A growing body of research explores geometric priors in visual recognition. For instance, CMT \cite{guo2022cmt} integrates convolutional operations into the Transformer backbone, introducing local receptive fields and translation equivariance as geometric priors for fine-grained feature modeling. PVTv2 \cite{wang2022pvt} employs overlapping convolutional patch embeddings and pyramid structures to encode multi-scale geometric priors, achieving strong adaptability in classification and segmentation. In addition, ConvNeXt \cite{liu2022convnet} demonstrates that carefully designed convolutional inductive biases can rival or even surpass Transformer architectures in vision benchmarks.  

In sequence modeling, ALiBi \cite{press2021train} replaces traditional positional embeddings with linear biases that depend on token distance, serving as a geometric prior to capture relative spatial regularities. This design enables effective extrapolation to longer sequences during inference. Similarly, relative positional encoding \cite{shaw2018self} has been widely adopted in both natural language and vision Transformers to improve structural consistency.  

For high-resolution vision tasks, geometric priors have also proven effective. For example, BPT \cite{sun2025ultra} introduces boundary-enhanced patch merging with explicit geometric constraints, adapting patch sizes according to object scales while emphasizing boundary features. Such designs strike a balance between global semantics and local geometric structures, yielding superior segmentation performance. In medical imaging, hybrid designs that combine CNN-based locality and Transformer-based global modeling \cite{chen2021transunet, xie2021cotr} have demonstrated the benefit of geometric priors in capturing multi-scale structural patterns.  

In summary, geometric priors can be incorporated in various ways—through convolutional locality, multi-scale pyramid designs, relative positional biases, or boundary-aware modeling. These priors enhance the structural sensitivity of deep networks and are especially valuable in medical imaging tasks, where domain-specific structures play a critical role in diagnosis.

\section{Materials and methods}

\begin{figure}[!htbp]
    \centering
    \includegraphics[width=\linewidth]{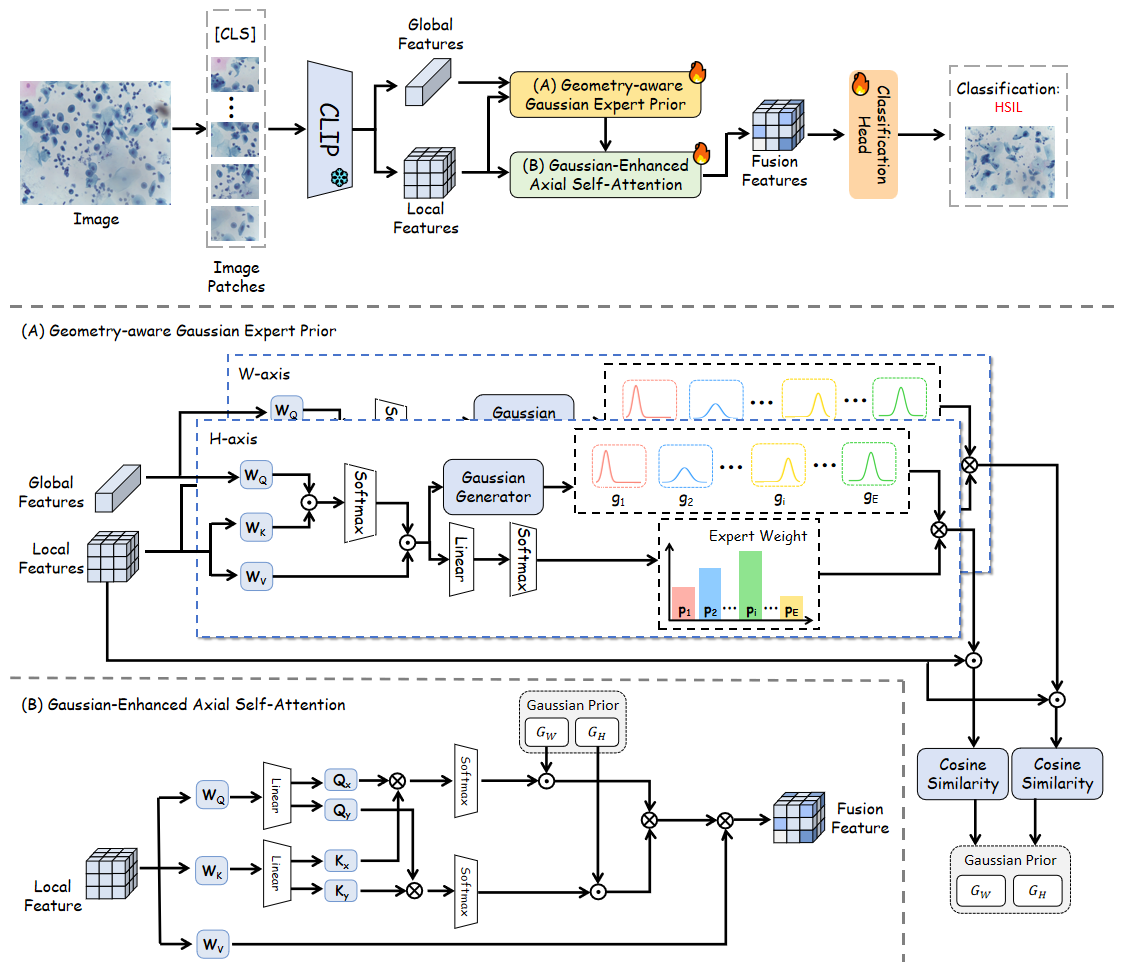} 
    \caption{
        Overall architecture of the proposed geometry-aware cervical cancer screening-oriented cytology classification framework.
        The input image is divided into patches and encoded by a pretrained CLIP backbone to obtain global and local features. 
        (A) The Geometry-aware Gaussian Expert Prior module generates direction-sensitive Gaussian priors by routing among multiple Gaussian experts conditioned on global semantics. 
        (B) The Gaussian-Enhanced Axial Self-Attention module injects these priors into axial attention along horizontal and vertical directions to produce fusion features, which are finally fed into the classification head.
    }
    \label{fig:framework}
\end{figure}

In this section, we introduce our proposed framework for cytology image classification, which integrates semantic abstraction with geometry-aware priors to achieve structurally consistent and interpretable predictions. The overall architecture is designed as a progressive pipeline that first extracts semantically enriched features, then constructs geometry-aware priors through Gaussian experts, and subsequently embeds these priors into an axial self-attention mechanism before final classification. In contrast to conventional CNN- or Transformer-based methods that rely purely on data-driven learning, our approach explicitly introduces structural constraints, providing both improved robustness and interpretability in medical applications. The framework is composed of four main components: (1) feature extraction, (2) geometry-aware prior generation, (3) Gaussian-enhanced axial self-attention, and (4) classification prediction. In what follows, we present each module in detail. 

An overview of the proposed framework is illustrated in Fig.~\ref{fig:framework}.

\subsection{Feature Extraction}
Cytology images typically exhibit substantial variations in cell morphology and spatial distribution, making it crucial to obtain features that combine both local and global semantics.  
Given an input cytology image $\mathbf{I} \in \mathbb{R}^{H \times W \times d}$, 
the image is partitioned into $P \times P$ non-overlapping patches, where each patch has spatial size $\tfrac{H}{P} \times \tfrac{W}{P}$.  
This produces a sequence of $N = P \cdot P$ patch embeddings:
\begin{equation}
    \mathbf{X} = [\mathbf{x}_1, \dots, \mathbf{x}_N] \in \mathbb{R}^{N \times d},
\end{equation}
where $\mathbf{x}_i$ denotes the $i$-th patch token and $d$ is the embedding dimension.  
This partitioning strategy ensures that each token summarizes a larger region of the cytology image, thereby reducing computational complexity while retaining sufficient structural information for downstream modeling.  

A learnable CLS token $\mathbf{q} \in \mathbb{R}^{d}$ is prepended to the patch sequence to encode global semantics and aggregate contextual information across the entire image.  
Subsequently, a hierarchical backbone $f_\theta(\cdot)$, initialized from a pretrained visual encoder (e.g., CLIP), 
processes the tokens and reshapes them into a 2D feature map:
\begin{equation}
    \mathbf{Z} = f_\theta([\mathbf{q}; \mathbf{X}]), 
    \quad \mathbf{Z} \in \mathbb{R}^{P \times P \times d}.
\end{equation}
Here, $\mathbf{q}$ captures high-level semantics, while $\mathbf{Z}$ retains spatially organized features with resolution $P \times P$.  
This dual representation ensures that the subsequent modules can exploit both global semantic abstraction and localized cell-level structures for enhanced interpretability.

\subsection{Geometry-aware Prior via Gaussian Experts}

While feature extraction provides meaningful embeddings, conventional attention mechanisms still rely solely on token-wise similarity, ignoring the inherent spatial organization present in cytology images. This limitation is particularly critical in medical imaging, where cells often exhibit structured or clinically relevant spatial patterns. To address this issue, we introduce a set of \emph{Gaussian experts} to produce \emph{directional Gaussian distributions} along spatial dimensions, and further construct geometry-aware priors to modulate subsequent axial attention.

\paragraph{Directional global descriptors.}
Let $\mathbf{X}\in\mathbb{R}^{H\times W\times D}$ be the local feature map and $\mathbf{c}\in\mathbb{R}^{512}$ be the global semantic token.We first obtain two direction-specific queries by linear projections:
\begin{equation}
\mathbf{q}_W = \mathbf{W}^{(W)}\mathbf{c},\qquad
\mathbf{q}_H = \mathbf{W}^{(H)}\mathbf{c},
\end{equation}
which correspond to the width- and height-direction Gaussian generators, respectively.
For each direction, we compute an enriched global descriptor via query-to-token cross-attention:
\begin{equation}
\mathbf{u}_d = \mathrm{Attn}(\mathbf{q}_d,\mathbf{X},\mathbf{X}),\qquad d\in\{W,H\},
\end{equation}
where $\mathbf{u}_d\in\mathbb{R}^{D}$ is shared to drive both expert routing and Gaussian parameter prediction.

\paragraph{Expert routing (Top-$K$) and Gaussian parameter prediction.}
For each direction, a semantic-guided router produces the activation probability over $E$ Gaussian experts:
\begin{equation}
\boldsymbol{\pi}_d = \mathrm{Softmax}(\mathbf{W}_r \mathbf{u}_d + \mathbf{b}_r),\qquad
\boldsymbol{\pi}_d \in \mathbb{R}^{E}.
\end{equation}
To encourage sparsity, we keep only Top-$K$ experts and renormalize their probabilities:
\begin{equation}
\pi'_{d,i}=\frac{\pi_{d,i}}{\sum_{j\in \mathrm{TopK}}\pi_{d,j}},\qquad i\in \mathrm{TopK}.
\end{equation}

Meanwhile, a Gaussian-parameter predictor $\psi(\cdot)$ outputs an offset and a width for each expert:
\begin{equation}
[\delta_{d,i},\, w_{d,i}] = \psi(\mathbf{u}_d),\qquad i=1,\dots,E.
\end{equation}
We apply bounded mappings in implementation (e.g., $\tanh$ for offsets and $\sigma(\cdot)$ for widths) to keep parameters in valid ranges, and combine them with a fixed reference grid $\{c_i\}_{i=1}^{E}$ to form each expert's mean and variance:
\begin{equation}
\mu_{d,i}=\mathrm{Clamp}(c_i+\delta_{d,i},0,1),\qquad
\sigma_{d,i}=\frac{\max(w_{d,i},\sigma_{\min})}{\gamma}.
\end{equation}
Accordingly, the $i$-th Gaussian expert along the normalized discrete axis $t\in[0,1]$ is:
\begin{equation}
g_{d,i}(t)=\frac{1}{\sigma_{d,i}\sqrt{2\pi}}\exp\!\left(-\frac{(t-\mu_{d,i})^2}{2\sigma_{d,i}^2}\right).
\end{equation}

\paragraph{Directional Gaussian distributions via router-weighted aggregation.}
To avoid wider kernels dominating the magnitude, we normalize each expert curve by its maximum response:
\begin{equation}
\tilde{g}_{d,i}(t)=\frac{g_{d,i}(t)}{\max_{t} g_{d,i}(t)}.
\end{equation}
The final \emph{directional Gaussian distribution} is obtained by router-weighted aggregation over the Top-$K$ experts:
\begin{equation}
\mathcal{G}_d(t)=\sum_{i\in\mathrm{TopK}}\pi'_{d,i}\,\tilde{g}_{d,i}(t),\qquad d\in\{W,H\}.
\end{equation}
In practice, $\mathcal{G}_d$ is generated on the discrete token grid of length $T=HW$, and reshaped back to the spatial map size for reweighting.This procedure is performed independently along the two axes, yielding $\mathcal{G}_W$ and $\mathcal{G}_H$.

\paragraph{Direction-aware feature reweighting.}
These directional Gaussian distributions are applied to the local feature map to obtain direction-aware weighted features:
\begin{equation}
\mathbf{X}_W=\mathcal{G}_W\odot \mathbf{X},\qquad
\mathbf{X}_H=\mathcal{G}_H\odot \mathbf{X},
\end{equation}
where $\odot$ denotes element-wise multiplication and Gaussian weights are broadcast along the channel dimension.

\paragraph{Construction of geometry-aware priors.}
Based on the direction-weighted features, we explicitly model structural relationships between spatial positions along each axis. Concretely, we build geometry-aware prior matrices using pairwise cosine similarity:
\begin{equation}
G_W(i,j)=\frac{\langle \mathbf{X}_W(i),\,\mathbf{X}_W(j)\rangle}{\|\mathbf{X}_W(i)\|\,\|\mathbf{X}_W(j)\|},\quad
i,j\in\{1,\dots,H\},
\end{equation}
\begin{equation}
G_H(i,j)=\frac{\langle \mathbf{X}_H(i),\,\mathbf{X}_H(j)\rangle}{\|\mathbf{X}_H(i)\|\,\|\mathbf{X}_H(j)\|},\quad
i,j\in\{1,\dots,W\}.
\end{equation}
Here, $\mathbf{X}_W(i)\in\mathbb{R}^{WD}$ denotes the feature vector of the $i$-th row after flattening the width and channel dimensions (consistent with the implementation that compares row-to-row relationships), and $\mathbf{X}_H(i)\in\mathbb{R}^{HD}$ is defined analogously for column-wise comparisons.

The resulting priors $G_W$ and $G_H$ encode structural consistency along the vertical and horizontal directions, respectively, and are used to modulate subsequent axial attention weights. By incorporating these geometry-aware priors, the attention mechanism is guided not only by semantic similarity but also by spatial organization inherent in cytological images, yielding more structurally consistent attention modeling.

\subsection{Gaussian-Enhanced Axial Self-Attention}
Building upon the geometry-aware priors, we propose a Gaussian-enhanced axial self-attention mechanism that jointly captures long-range dependencies and fine-grained structural relations. While standard self-attention is effective for global context modeling, its quadratic complexity with respect to spatial size renders it inefficient for high-resolution cytology images, and it lacks explicit encoding of spatial plausibility. Our method addresses these limitations by decomposing the 2D attention into two one-dimensional operations and incorporating Gaussian priors, thereby achieving efficiency while embedding structural awareness.

\paragraph{Standard axial attention.}  
Given an input feature map $\mathbf{Z} \in \mathbb{R}^{P \times P \times d}$, query, key, and value projections are obtained as:
\begin{equation}
\mathbf{Q} = \mathbf{Z}\mathbf{W}_Q, \quad
\mathbf{K} = \mathbf{Z}\mathbf{W}_K, \quad
\mathbf{V} = \mathbf{Z}\mathbf{W}_V.
\end{equation}
The attention is then computed separately along each axis. For the width-wise direction:
\begin{equation}
\tilde{\alpha}^{(W)}_{ij} = 
\mathrm{Softmax}_j \!\left(
\frac{\mathbf{Q}^{(W)}_i \cdot \mathbf{K}^{(W)}_j}{\sqrt{d}}
\right),
\end{equation}
where $i,j \in \{1,\dots,P\}$ index different columns.  
Analogously, for the height-wise direction:
\begin{equation}
\tilde{\alpha}^{(H)}_{ij} = 
\mathrm{Softmax}_j \!\left(
\frac{\mathbf{Q}^{(H)}_i \cdot \mathbf{K}^{(H)}_j}{\sqrt{d}}
\right),
\end{equation}
where $i,j \in \{1,\dots,P\}$.  
This axial decomposition reduces the computational complexity from $\mathcal{O}((HW)^2)$ to:
\[
\mathcal{O}\left(P^2 \cdot P + P \cdot P^2\right) = \mathcal{O}(2P^3),
\]
making the mechanism significantly more scalable for cytology image analysis.  

\paragraph{Incorporating geometry-aware priors.}  
To further encode spatial plausibility, Gaussian priors $\{G^{(W)}, G^{(H)}\}$ are introduced as multiplicative modulation terms. For the width-wise direction:
\begin{equation}
\alpha^{(W)}_{ij} = \tilde{\alpha}^{(W)}_{ij} \cdot \exp(-\gamma \, G^{(W)}_{ij}),
\end{equation}
and similarly for the height-wise direction:
\begin{equation}
\alpha^{(H)}_{ij} = \tilde{\alpha}^{(H)}_{ij} \cdot \exp(-\gamma \, G^{(H)}_{ij}),
\end{equation}
where $\gamma$ is a decay coefficient controlling the influence of the priors.  
This modulation encourages attention to focus on geometrically consistent regions while suppressing implausible correlations.

\paragraph{Feature update and fusion.}
With the prior-modulated attention weights, the final feature representation is obtained by sequential width–height aggregation:
\begin{equation}
    \mathbf{Z}' = \alpha^{(H)} \alpha^{(W)} \mathbf{V}, 
    \quad 
    \mathbf{Z}' \in \mathbb{R}^{P \times P \times d}.
\end{equation}

This design preserves the global context modeling power of attention while embedding explicit structural awareness, ensuring sensitivity to subtle morphological variations that are critical in cytological diagnosis.

\subsection{Classification Prediction}
Finally, the enriched representation $\mathbf{Z}' \in \mathbb{R}^{P \times P \times d}$ is transformed into categorical predictions via a lightweight classification head. The objective of this step is to map the structurally consistent embeddings into the decision space while maintaining computational efficiency.  

First, the spatial feature map is aggregated into a compact vector using global average pooling:
\begin{equation}
    \mathbf{h} = \mathrm{GAP}(\mathbf{Z}'), \quad \mathbf{h} \in \mathbb{R}^{d}.
\end{equation}
This operation ensures that the pooled vector encodes the holistic representation of the image by averaging over all spatial locations.  

Then, the pooled feature vector is mapped into the label space through a fully connected transformation:
\begin{equation}
    \mathbf{o} = \mathrm{FC}(\mathbf{h}) = \mathbf{W}_{\text{cls}} \mathbf{h} + \mathbf{b}_{\text{cls}}, \quad \mathbf{o} \in \mathbb{R}^{C},
\end{equation}
where $C$ denotes the number of classes and $\mathbf{W}_{\text{cls}}, \mathbf{b}_{\text{cls}}$ are trainable parameters.  

Finally, the class distribution is obtained via:
\begin{equation}
    \hat{\mathbf{y}} = \mathrm{Softmax}(\mathbf{o}).
\end{equation}

For training, we employ the standard cross-entropy loss function:
\begin{equation}
    \mathcal{L}_{\text{cls}} = - \sum_{c=1}^{C} y_c \log \hat{y}_c,
\end{equation}
where $y_c$ is the one-hot ground-truth label and $\hat{y}_c$ is the predicted probability for class $c$.  
This objective encourages the model to maximize the likelihood of the correct class, aligning the final predictions with clinically validated annotations.

\section{Experiment Settings}
\subsection{Datasets}

We evaluate the proposed method on two publicly available cervical cytology datasets commonly used for automated cervical cancer screening-related image classification.

\textbf{Mendeley LBC} \cite{HUSSAIN2020105589} includes 963 high-resolution images annotated with four diagnostic categories: Normal (NL), Low-Grade Squamous Intraepithelial Lesions (LSIL), High-Grade Squamous Intraepithelial Lesions (HSIL), and Squamous Cell Carcinoma (SCC). Each image has a resolution of $2048 \times 1536$ pixels. The dataset is derived from real clinical samples and is similarly divided into 80\% training and 20\% testing sets.

\textbf{SIPaKMeD} \cite{8451588} consists of 4049 single-cell images categorized into five classes: Superficial-Intermediate, Parabasal, Koilocytotic, Dyskeratotic, and Metaplastic. Each class contains approximately 800 images, resulting in a balanced class distribution. All images are of size $256 \times 256$ pixels. The dataset is split into 80\% for training and 20\% for testing.

\begin{table}[t]
\centering
\caption{Arrangement of the cervical cytology datasets.}
\begin{tabular}{l l c}
\toprule
Dataset & Category & No.\ of Images \\
\midrule
\multirow{4}{*}{Mendeley LBC} 
    & NL   & 613 \\
    & LSIL & 113 \\
    & HSIL & 163 \\
    & SCC  & 74 \\
\midrule
\multirow{5}{*}{SIPaKMeD}
    & Dyskeratotic          & 813 \\
    & Metaplastic           & 793 \\
    & Koilocytotic          & 825 \\
    & Parabasal             & 787 \\
    & Superficial-Intermediate & 831 \\
\bottomrule
\end{tabular}
\label{tab:dataset_arrangement} 
\end{table}

\subsection{Metrics}

To comprehensively assess the model’s performance in multi-class classification, we adopt Accuracy, Precision, Recall, and F1-score as evaluation metrics. Except for Accuracy, which measures the overall proportion of correctly classified samples, the other metrics are computed using macro-averaging. This approach calculates each metric independently for each class and then averages across all classes, ensuring that all categories contribute equally to the final evaluation. Precision and Recall evaluate the model’s ability to correctly identify and recover each class, while the F1-score balances both to provide a robust measure of classification effectiveness, particularly in the presence of class imbalance.

Suppose the classification task involves $C$ classes and the total number of validation samples is $N$. For any sample $j=1,2,...,N$, let the ground truth label be $y_j \in \{1, 2,...,C\}$, and the predicted label be $\hat{y}_j \in \{1,2,...,C\}$. For each class $i$, define: $TP_i$ is the number of true positives correctly classified as class $i$; $FP_i$ is the number of false positives incorrectly predicted as class $i$; $FN_i$ is the number of false negatives where true class is $i$ but predicted otherwise.

\begin{align}
&\text{Precision} = \frac{1}{C} \sum_{i=1}^{C} \frac{TP_i}{TP_i + FP_i}, \\
&\text{Recall} = \frac{1}{C} \sum_{i=1}^{C} \frac{TP_i}{TP_i + FN_i}, \\
&\text{F1} = \frac{1}{C} \sum_{i=1}^{C} \frac{2 \cdot TP_i}{2 \cdot TP_i + FP_i + FN_i}, \\
&\text{ACC} = \frac{\sum_{i=1}^{C} TP_i}{\sum_{i=1}^{C} (TP_i + FN_i)} 
= \frac{\sum_{i=1}^{C} TP_i}{N}.
\end{align}

\subsection{Implementation Details}
Our framework is implemented in PyTorch and trained on a single NVIDIA A40 GPU. 
The input cytology images are resized to $224 \times 224$. 
Each image is divided into $16 \times 16$ non-overlapping patches, resulting in $196$ patch tokens. 
An additional learnable CLS token is appended, forming a sequence of $197$ tokens. 
Each token is embedded into a $768$-dimensional space using the CLIP-ViT-B/16 encoder~\cite{radford2021learning}, which serves as the feature backbone.  

For geometry-aware prior modeling, we employ a Gaussian Mixture-of-Experts (MoE) module inspired by QA-TIGER~\cite{kim2025question}.  
The module contains $10$ independent experts, among which the Top-$K$ experts (with $K=8$ for LBC and $K=6$ for SIPaKMeD) are dynamically selected for each input via a semantic-guided router.  
Each expert is implemented as a two-layer MLP with ReLU activation, predicting Gaussian centers and widths using $\tanh$ and $\mathrm{sigmoid}$ constraints, respectively.  
A Gaussian scale normalization factor of $9$ and a dropout rate of $0.1$ are adopted.  

The decoder is constructed upon the DFormerV2 architecture~\cite{yin2025dformerv2}, consisting of three hierarchical stages with embedding dimensions $\{512, 1024, 2048\}$. 
Each stage contains $\{4, 8, 8\}$ Transformer blocks with attention heads $\{4, 8, 8\}$ and attention ranges $\{4, 6, 6\}$. 
Feed-forward layers adopt expansion ratios of $\{4, 4, 3\}$. 
Layer normalization is applied throughout, while LayerScale is disabled. 
A stochastic depth strategy is employed with a maximum drop path rate of $0.25$, and hierarchical Patch Merging is performed in the first two stages.  

For classification, a single fully connected layer is used after global average pooling. 
Training is performed with the Cross-Entropy Loss function. 
The learning rate follows a warm-up strategy: it increases linearly during the first 20 epochs for LBC and 40 epochs for SIPaKMeD. 
The peak learning rate is set to $5\times10^{-6}$ for LBC and $1\times10^{-5}$ for SIPaKMeD. 
Both models are optimized for $100$ epochs with momentum $0.9$, weight decay $0.01$, batch size $16$, and polynomial decay factor $0.9$.

\section{Results and Discussion}

\subsection{Comparisons with Other Methods}

\begin{table}[t]
\renewcommand{\arraystretch}{1.15}
\centering
\caption{Comparison with other methods on two challenging datasets. 
The best results are highlighted in bold. The second-best results are marked by underline.}
\label{tab:comparison}
\scriptsize
\resizebox{\textwidth}{!}{%
\begin{tabular}{lcccc|lcccc}
\toprule
\multicolumn{5}{c|}{\textbf{Mendeley LBC}} & \multicolumn{5}{c}{\textbf{SIPaKMeD}} \\
\cmidrule(lr){1-5} \cmidrule(lr){6-10}
Methods & Acc & Recall & Precision & F1 & Methods & Acc & Recall & Precision & F1 \\
\midrule
GoogleNet \cite{Szegedy_2015_CVPR}        & 95.40 & 95.42 & 95.52 & 95.47 
& VGG16 \cite{simonyan2014very}              & 83.36 & 83.65 & 86.16 & 84.89 \\
ResNet50 \cite{He_2016_CVPR}              & 96.42 & 94.44 & 96.42 & 95.42 
& GoogleNet \cite{Szegedy_2015_CVPR}         & 84.03 & 84.31 & 86.60 & 85.44 \\
VGG16 \cite{simonyan2014very}              & 97.16 & 97.17 & 97.20 & 97.18 
& DenseNet169 \cite{huang2017densely}        & 84.69 & 84.97 & 87.05 & 86.00 \\
DenseNet121 \cite{huang2017densely}        & 97.34 & 97.32 & 97.36 & 97.34 
& ResNet50 \cite{He_2016_CVPR}              & 84.84 & 85.12 & 86.77 & 85.94 \\
Bilal et al. \cite{bilal2025automated}     & 97.94 & 94.80 & 97.48 & 96.03 
& Haryanto et al. \cite{haryanto2020utilization} & 84.88 & 84.88 & 89.18 & 85.35 \\
Idlahcen et al. \cite{idlahcen2025pathocoder} & 98.37 & 98.37 & 98.41 & 98.37 
& EfficientNet-B0 \cite{tan2019efficientnet} & 87.48 & 87.68 & 88.25 & 87.96 \\
Khowaja et al. \cite{khowaja2024enhancing} & 98.44 & 97.99 & 97.34 & 97.66 
& Niswati et al. \cite{niswati2021perbandingan} & 91.00 & 91.00 & 91.00 & 91.00 \\
EfficientNet-B0 \cite{tan2019efficientnet} & 98.51 & 98.53 & 98.53 & 98.53 
& Alsubai et al. \cite{alsubai2023privacy}   & 91.13 & 91.14 & 91.37 & 91.20 \\
Bilal et al. \cite{BILAL2024112366}        & 98.96 & 98.00 & 98.75 & 98.50 
& Shaik et al. \cite{shaik2025deep}          & 91.28 & 91.00 & 91.00 & 91.00 \\
Xu et al. \cite{xu2025dualbranch}          & 99.07 & 99.09 & 99.08 & 99.08 
& Wubineh et al. \cite{wubineh2024classification} & 92.00 & 92.00 & 92.00 & 92.00 \\
Sahoo et al. \cite{sahoo2023enhancing}     & 99.22 & 99.17 & 99.22 & 99.19 
& Pacal et al. \cite{pacal2023deep}          & 92.95 & 92.71 & 93.89 & 93.30 \\
Manna et al. \cite{manna2021fuzzy}          & 99.23 & \underline{99.23} & 99.13 & 99.18 
& Comert et al. \cite{comert2025convolutional} & 92.84 & 92.84 & 93.06 & 92.82 \\
Kaur et al. \cite{kaur2022mlnet}            & \underline{99.36} & \textbf{99.32} & \underline{99.35} & \textbf{99.32} 
& Tripathi et al. \cite{tripathi2021classification} & \underline{94.89} & \underline{94.93} & \textbf{95.78} & \underline{95.29} \\
\textbf{Proposed Method}                    & \textbf{99.48} & {99.22} & \textbf{99.48} & \underline{99.27} 
& \textbf{Proposed Method}                   & \textbf{96.08} & \textbf{95.99} & \underline{95.62} & \textbf{95.40} \\
\bottomrule
\end{tabular}%
}
\end{table}

\begin{figure}[t]
    \centering
    \includegraphics[width=0.9\linewidth]{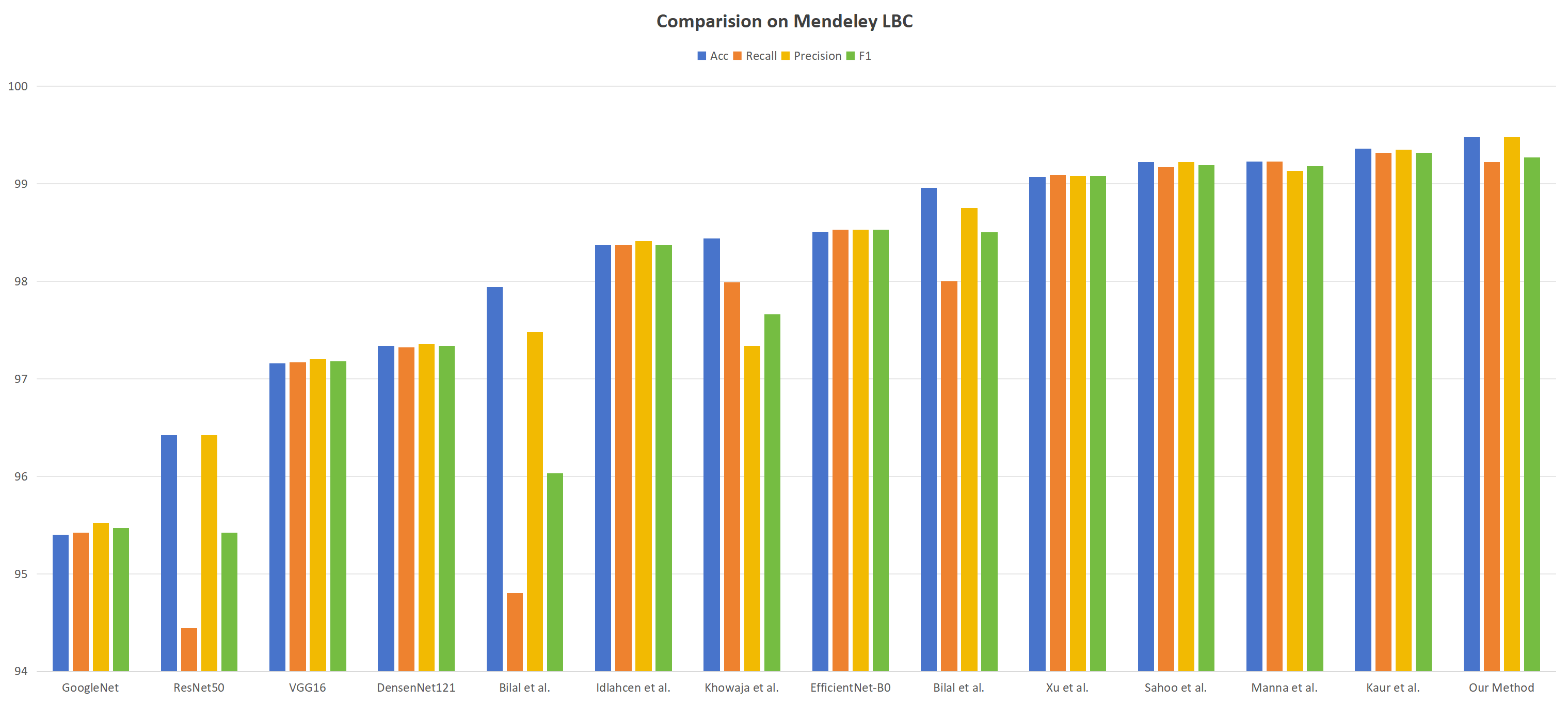}
    \caption{Comparison with baseline methods on the Mendeley LBC dataset.}
    \label{fig:lbc_comparison}
\end{figure}

To comprehensively evaluate the effectiveness of the proposed framework, we compare it with a wide range of recent methods on the Mendeley LBC and SIPaKMeD datasets, as summarized in \autoref{tab:comparison}.

On the Mendeley LBC dataset, the proposed method achieves the highest classification accuracy of \textbf{99.48\%}, outperforming recent strong approaches such as Bilal et al.~\cite{BILAL2024112366} (98.96\%) and Xu et al.~\cite{xu2025dualbranch} (99.07\%) by margins of 0.52\% and 0.41\%, respectively.
In terms of precision, our method reaches \textbf{99.48\%}, which is the best among all compared methods, while achieving an F1-score of 99.27\%, ranking second overall.
Although Kaur et al.~\cite{kaur2022mlnet} reports a slightly higher recall (99.32\% vs.\ 99.22\%), our framework demonstrates a more balanced performance across all four metrics, avoiding the trade-off between sensitivity and precision that is commonly observed in prior methods.

The comparative performance trends across different baselines are further illustrated in Fig.~\ref{fig:lbc_comparison}, where the proposed method consistently ranks at the top across accuracy, recall, precision, and F1-score.

On the more challenging SIPaKMeD dataset, which exhibits higher inter-class similarity and morphological diversity, the proposed method also achieves the best overall performance.
Specifically, it attains an accuracy of \textbf{96.08\%}, surpassing recent competitive methods such as Tripathi et al.~\cite{tripathi2021classification} (94.89\%) and Pacal et al.~\cite{pacal2023deep} (92.95\%) by 1.19\% and 3.13\%, respectively.
Moreover, our approach yields the highest recall of \textbf{95.99\%} and the best F1-score of \textbf{95.40\%}, indicating superior sensitivity to abnormal cell patterns while maintaining robust overall classification quality.

\begin{figure}[t]
    \centering
    \includegraphics[width=0.9\linewidth]{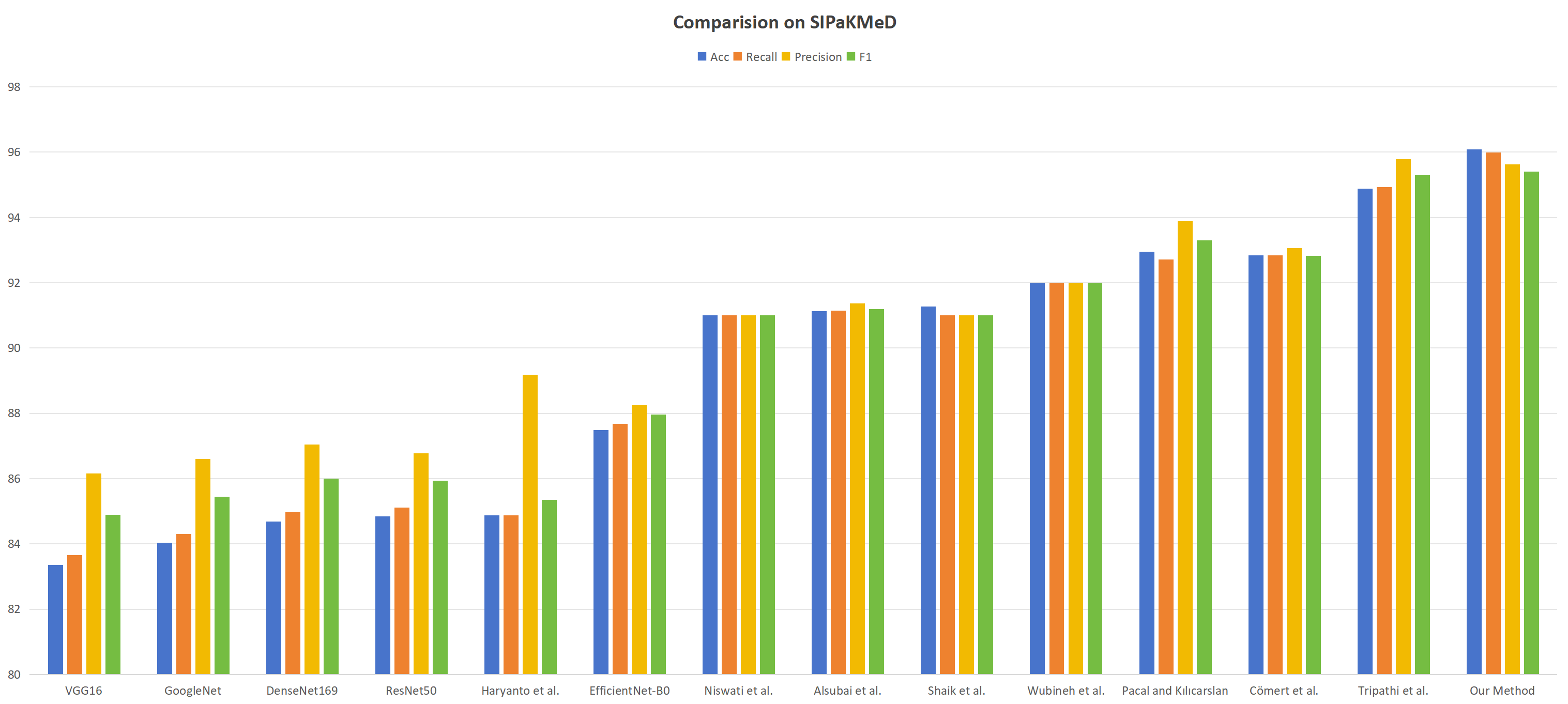}
    \caption{Comparison with baseline methods on the SIPaKMeD dataset.}
    \label{fig:sipakmed_comparison}
\end{figure}

As shown in Fig.~\ref{fig:sipakmed_comparison}, the proposed framework consistently outperforms conventional CNN-based and hybrid methods across all evaluation metrics, highlighting its effectiveness in handling visually ambiguous cytological categories.

\subsection{Analysis of the Proposed Method}

To further analyze the behavior of the proposed framework beyond aggregate metrics, we provide a detailed evaluation using confusion matrices, ROC curves, and qualitative confidence visualizations.

\begin{figure}[t]
    \centering
    \includegraphics[width=0.9\linewidth]{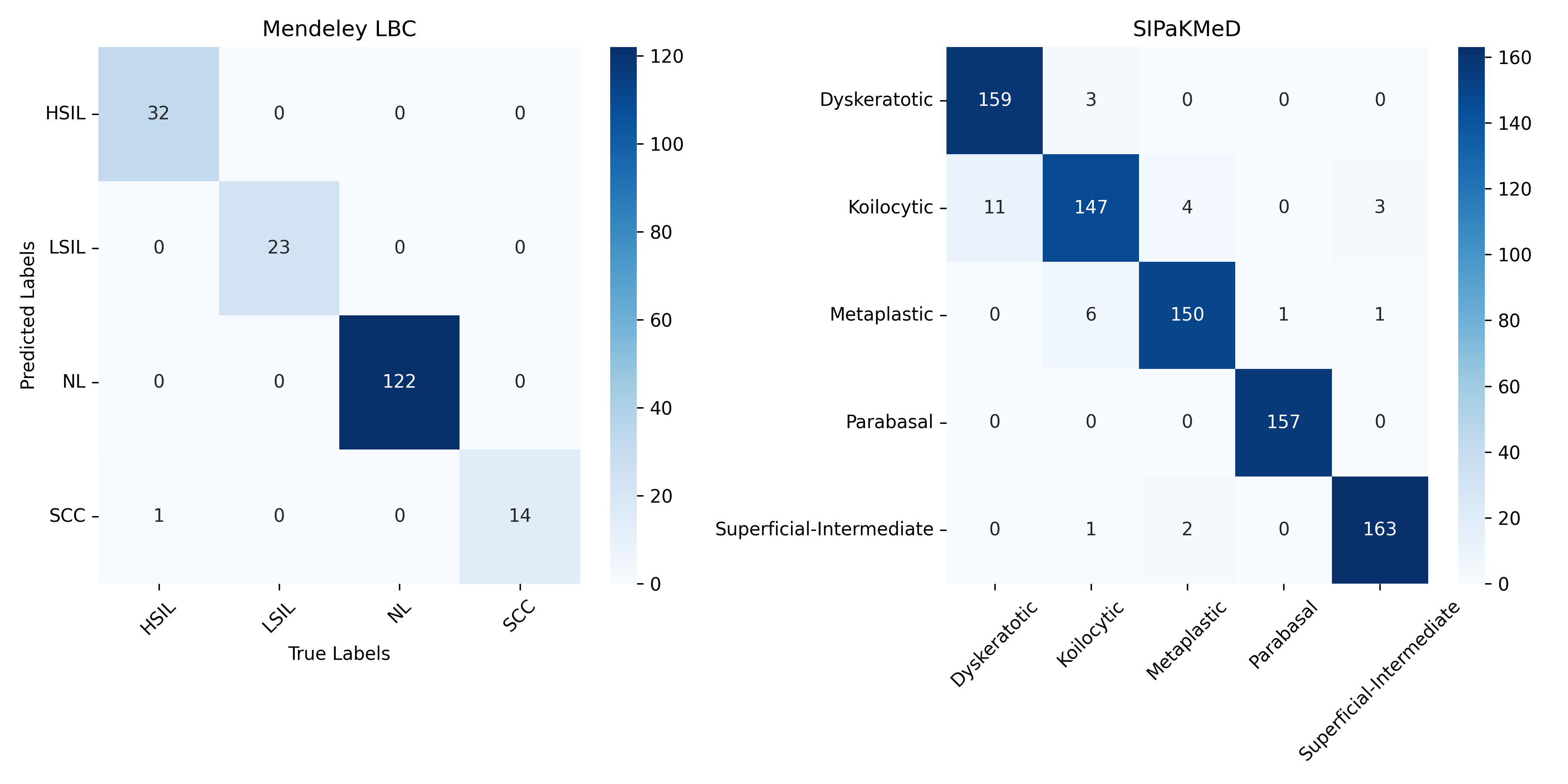}
    \caption{Confusion matrices of the proposed method on the Mendeley LBC and SIPaKMeD datasets.}
    \label{fig:confusion_matrices}
\end{figure}

The confusion matrices in Fig.~\ref{fig:confusion_matrices} provide a fine-grained view of the classification behavior of the proposed method.
On the Mendeley LBC dataset, the majority of samples are correctly classified, with only one misclassification observed between SCC and HSIL, while all NL samples are perfectly recognized.
Similarly, on the SIPaKMeD dataset, most errors occur between Koilocytotic and Dyskeratotic categories, which are known to exhibit highly similar morphological characteristics.
Overall, the diagonal dominance in both matrices confirms the strong discriminative capability and stability of the proposed model.

\begin{figure}[t]
  \centering
  \begin{minipage}{0.45\textwidth}
    \centering
    \includegraphics[width=\linewidth]{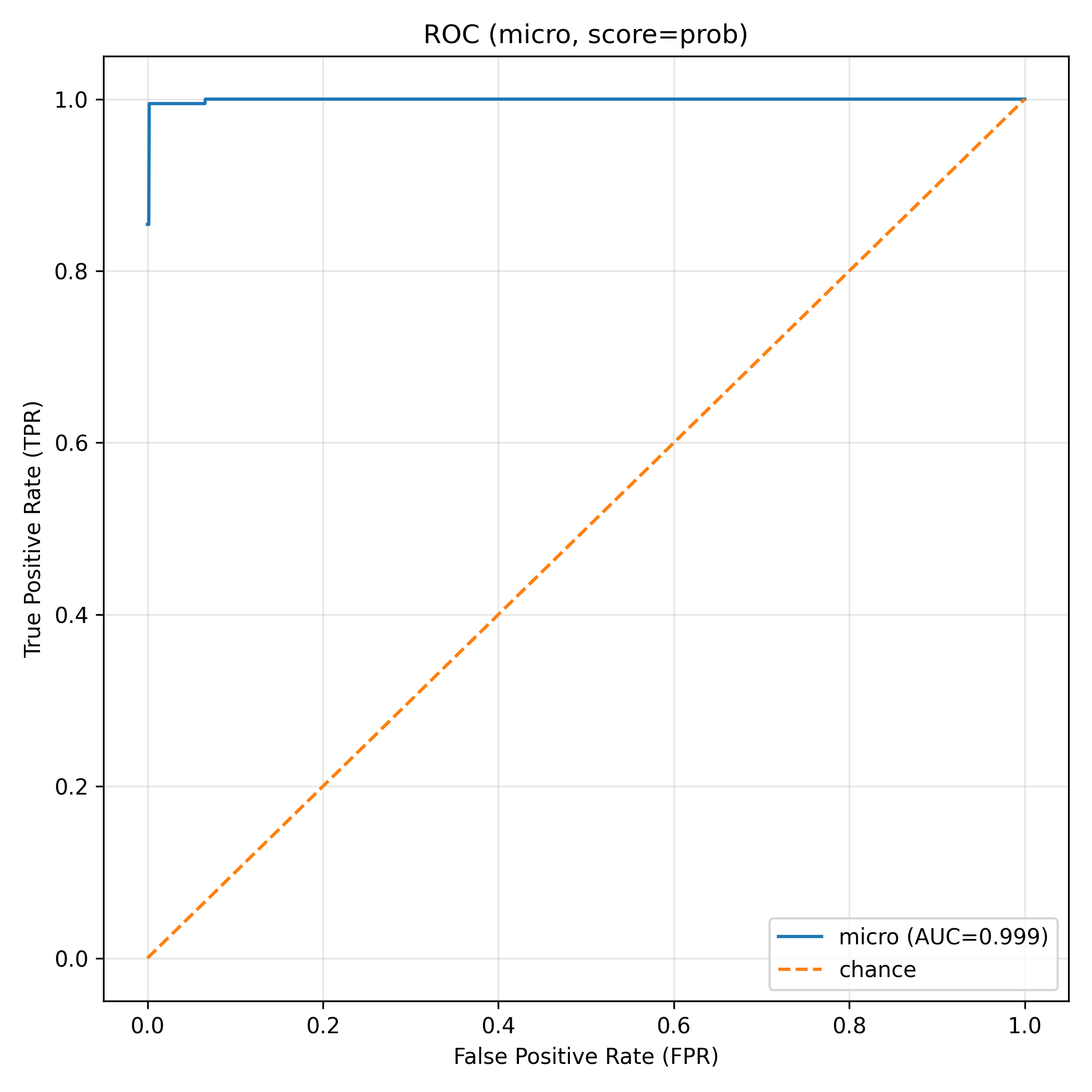}\\
    \textbf{(a)} Mendeley LBC
  \end{minipage}
  \hfill
  \begin{minipage}{0.45\textwidth}
    \centering
    \includegraphics[width=\linewidth]{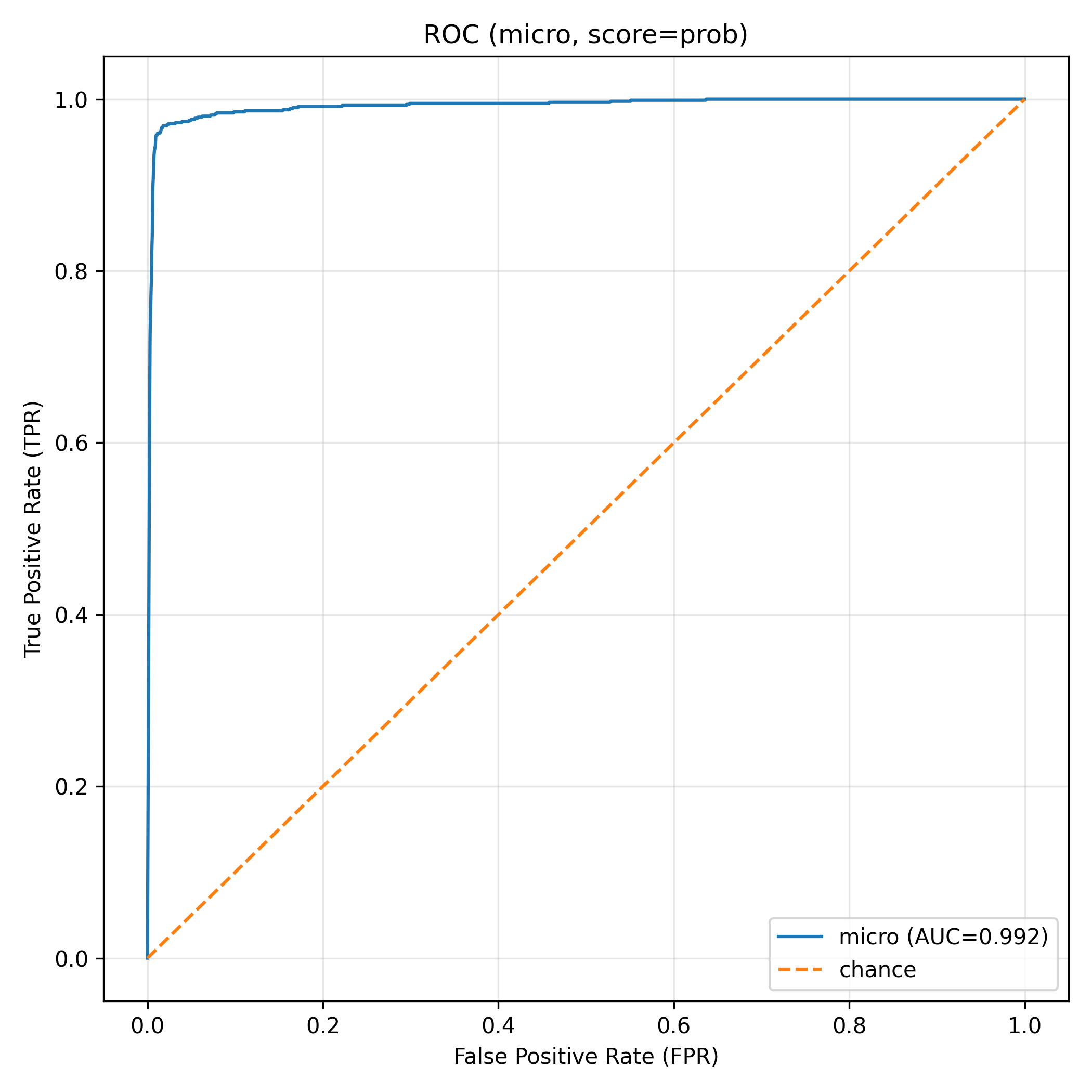}\\
    \textbf{(b)} SIPaKMeD
  \end{minipage}
  \caption{Micro-average ROC curves of the proposed method on two datasets: (a) Mendeley LBC and (b) SIPaKMeD.}
  \label{fig:roc_curves}
\end{figure}

To further assess the overall separability of different classes under varying decision thresholds, we report the micro-average ROC curves in Fig.~\ref{fig:roc_curves}.
The proposed method achieves a micro-AUC of approximately 0.999 on the Mendeley LBC dataset and 0.992 on the SIPaKMeD dataset, indicating near-perfect class separability.
These results demonstrate that the learned representations are not only accurate at a fixed operating point, but also remain robust across a wide range of confidence thresholds.

\begin{figure}[t]
    \centering
    \begin{minipage}{0.75\textwidth}
        \centering
        \includegraphics[width=\linewidth]{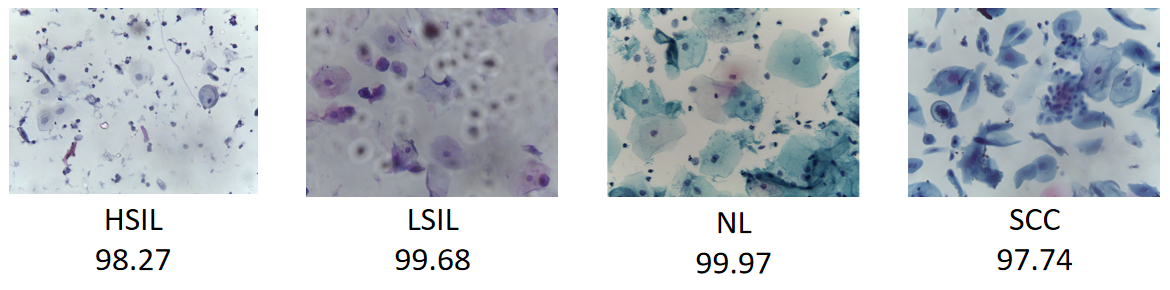}\\
        \textbf{(a)} Mendeley LBC dataset.
    \end{minipage}

    \vspace{0.6em}

    \begin{minipage}{0.95\textwidth}
        \centering
        \includegraphics[width=\linewidth]{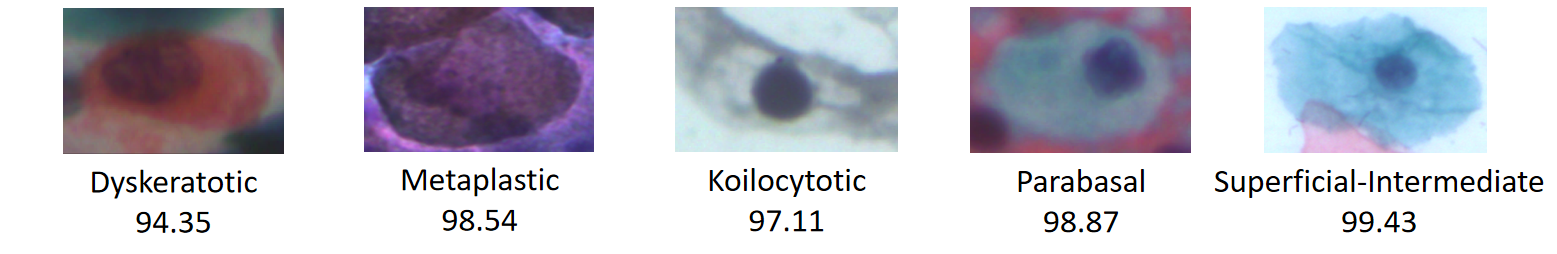}\\
        \textbf{(b)} SIPaKMeD dataset.
    \end{minipage}

    \caption{Visualization of class-wise prediction confidence (\%) on two cervical cytology datasets: (a) Mendeley LBC and (b) SIPaKMeD.}
    \label{fig:confidence_visualization}
\end{figure}

Figure~\ref{fig:confidence_visualization} presents a qualitative visualization of class-wise prediction confidence produced by the proposed method.
On the Mendeley LBC dataset (Fig.~\ref{fig:confidence_visualization}a), all four categories receive confidence scores exceeding 97\%, with the NL class reaching as high as 99.97\%.
Similarly, on the SIPaKMeD dataset (Fig.~\ref{fig:confidence_visualization}b), visually challenging categories such as Koilocytotic and Dyskeratotic are predicted with confidence scores of 97.11\% and 94.35\%, respectively.
These consistently high confidence values across both datasets indicate that the proposed geometry-aware framework yields stable and reliable predictions even under significant intra-class variability.

\subsection{Ablation Study}

\begin{table}[t]
\renewcommand{\arraystretch}{1.1} 
\centering
\small
\caption{Ablation study on the effect of Transformer encoder depth. Best results on each dataset are highlighted in bold.}
\label{tab:transformer_depth_ablation}
\begin{tabularx}{\textwidth}{c | *{4}{>{\centering\arraybackslash}X} | *{4}{>{\centering\arraybackslash}X}}
\toprule
\multirow{2}{*}{Layers} & \multicolumn{4}{c|}{Mendeley LBC} & \multicolumn{4}{c}{SIPaKMeD} \\
\cmidrule(lr){2-5} \cmidrule(lr){6-9}
 & Acc & Recall & Precision & F1 & Acc & Recall & Precision & F1 \\
\midrule
2 & 98.96 & 97.66 & 98.18 & 97.44 & 95.04 & 95.64 & 95.89 & 95.42 \\
3 & \textbf{99.48} & \textbf{99.22} & \textbf{99.48} & \textbf{99.27} & \textbf{96.08} & \textbf{95.99} & \textbf{95.62} & \textbf{95.40} \\
4 & 98.96 & 98.18 & 98.96 & 98.36 & 91.97 & 91.64 & 92.01 & 90.86 \\
\bottomrule
\end{tabularx}
\end{table}

\begin{table}[t]
\renewcommand{\arraystretch}{1.1} 
\centering
\small
\caption{Ablation study on the effectiveness of the geometric prior module. Best results on each dataset are highlighted in bold.}
\label{tab:geometric_prior_ablation}
\begin{tabularx}{\textwidth}{c | *{4}{>{\centering\arraybackslash}X} | *{4}{>{\centering\arraybackslash}X}}
\toprule
\multirow{2}{*}{Setting} & \multicolumn{4}{c|}{Mendeley LBC} & \multicolumn{4}{c}{SIPaKMeD} \\
\cmidrule(lr){2-5} \cmidrule(lr){6-9}
 & Acc & Recall & Precision & F1 & Acc & Recall & Precision & F1 \\
\midrule
w/o geometric prior & 98.44 & 97.55 & 97.29 & 96.91 & 94.55 & 95.16 & 94.10 & 93.98 \\
w/  geometric prior    & \textbf{99.48} & \textbf{99.22} & \textbf{99.48} & \textbf{99.27} & \textbf{96.08} & \textbf{95.99} & \textbf{95.62} & \textbf{95.40} \\
\bottomrule
\end{tabularx}
\end{table}

\begin{table}[t]
\renewcommand{\arraystretch}{1.1} 
\centering
\small
\caption{Performance with different numbers of experts in Gaussian MoE.}
\label{tab:exp_num}
\begin{tabularx}{\textwidth}{c | *{4}{>{\centering\arraybackslash}X} | *{4}{>{\centering\arraybackslash}X}}
\toprule
\multirow{2}{*}{topK} & \multicolumn{4}{c|}{Mendeley LBC} & \multicolumn{4}{c}{SIPaKMeD} \\
\cmidrule(lr){2-5} \cmidrule(lr){6-9}
 & Acc & Recall & Precision & F1 & Acc & Recall & Precision & F1 \\
\midrule
2  & 98.96 & 98.44 & 97.66 & 97.62 & 95.53 & 94.84 & 95.16 & 94.31  \\
4  & 98.96 & 98.85 & 96.88 & 97.29 & 95.47 & 95.51 & 95.18 & 94.66  \\
6  & 99.48 & \textbf{99.48} & 98.44 & 98.67 & \textbf{96.08} & \textbf{95.99} & \textbf{95.62} & \textbf{95.40}  \\
8  & \textbf{99.48} & 99.22 & \textbf{99.48} & \textbf{99.27} & 95.34 & 95.88 & 94.82 & 94.69  \\
10 & 99.48 & 98.96 & 99.38 & 99.03 & 94.55 & 94.20 & 94.97 & 93.94  \\
\bottomrule
\end{tabularx}
\end{table}

\begin{table}[t]
\renewcommand{\arraystretch}{1.1} 
\centering
\small
\caption{Ablation study on the effectiveness of different axes in the axial self-attention mechanism.}
\label{tab:axial_att}
\begin{tabularx}{\textwidth}{cc | *{4}{>{\centering\arraybackslash}X} | *{4}{>{\centering\arraybackslash}X}}
\toprule
\multicolumn{2}{c|}{Setting} & \multicolumn{4}{c|}{Mendeley LBC} & \multicolumn{4}{c}{SIPaKMeD} \\
\cmidrule{1-10}
H& W & Acc & Recall & Precision & F1 & Acc & Recall & Precision & F1 \\
\midrule
{$\times$}&{$\times$}  & 98.44 & 96.35 & 97.14 & 96.45 & 93.93 & 93.44 & 93.81 & 92.96 \\
{$\times$}&{$\checkmark$} & 98.96 & 97.92 & 98.96 & 97.81 & 94.79 & 94.44 & 94.61 & 94.01 \\
{$\checkmark$}&{$\times$}  & 98.96 & 98.18 & 98.75 & 98.15 & 95.22 & 94.60 & 94.85 & 94.04 \\
{$\checkmark$}&{$\checkmark$}  & \textbf{99.48} & \textbf{99.22} & \textbf{99.48} & \textbf{99.27} & \textbf{96.08} & \textbf{95.99} & \textbf{95.62} & \textbf{95.40} \\
\bottomrule
\end{tabularx}
\end{table}

To gain deeper insights into the contributions of individual modules and design choices, we performed a series of ablation studies while keeping the backbone architecture unchanged. Specifically, we systematically investigated four key factors: the depth of Transformer layers, the incorporation of the geometric prior module, the number of experts in the Gaussian Mixture-of-Experts mechanism, and the strategy adopted for the classification head. These studies allow us to isolate and quantify the effectiveness of each component, thereby clarifying their respective roles in enhancing the overall model performance.

\subsubsection{Effect of Transformer Depth}

To assess the influence of Transformer encoder depth on classification performance, we conducted an ablation study by varying the number of encoder layers (2, 3, and 4) on both the Mendeley LBC and SIPaKMeD datasets.
As summarized in \autoref{tab:transformer_depth_ablation}, performance improves steadily when increasing the encoder depth from 2 to 3 layers, but degrades when further extending to 4 layers.
Specifically, the 3-layer configuration achieves the best performance across all metrics, with 99.48\% accuracy and 99.27\% F1-score on Mendeley LBC, and 96.08\% accuracy with 95.40\% F1-score on SIPaKMeD.
In contrast, while the 2-layer model exhibits slightly weaker representation capability, the 4-layer configuration suffers from unstable optimization and gradient explosion, leading to a sharp performance drop (accuracy falling to 91.97\% on SIPaKMeD).
These results indicate that a moderate encoder depth (three layers) provides a good balance between representational power, numerical stability, and generalization, ensuring reliable convergence without excessive complexity.

\subsubsection{Effectiveness of Geometric Priors}

To verify the effectiveness of the proposed geometric prior module, we performed an ablation experiment by removing this component while keeping all other settings identical. The results, summarized in \autoref{tab:geometric_prior_ablation}, demonstrate that incorporating geometric priors leads to consistent and substantial performance gains on both datasets.
On the Mendeley LBC dataset, the model with geometric priors achieves a significant improvement in all metrics, particularly in F1-score, which rises from 96.91\% to 99.27\%. This highlights the strong contribution of spatially guided priors in capturing structural regularities among cervical cells.
Similarly, on the SIPaKMeD dataset, the inclusion of geometric priors enhances overall performance, boosting accuracy from 94.55\% to 96.08\% and F1-score from 93.98\% to 95.40\%.
These consistent improvements indicate that the proposed geometry-aware mechanism effectively complements semantic representations by introducing spatial constraints, thereby enhancing the model’s ability to distinguish morphologically similar cell types across different datasets.

\subsubsection{Impact of Expert Number in Gaussian MoE}

To assess the influence of the number of experts in the Gaussian Mixture-of-Experts (MoE) module, we conducted experiments by varying the Top-$K$ values while keeping the routing mechanism and other hyperparameters unchanged. The results, summarized in \autoref{tab:exp_num}, indicate that an appropriate number of experts plays a crucial role in balancing representation diversity and model stability.

On the \textbf{Mendeley LBC} dataset, the model maintains consistently high performance across all configurations. Both Top-6 and Top-8 settings achieve the highest accuracy of 99.48\%, with Top-8 yielding the best precision (99.48\%) and F1-score (99.27\%). This suggests that moderate expert diversity enables more effective specialization among Gaussian experts, leading to richer geometric priors and more discriminative representations. However, further increasing the number of experts to 10 introduces redundant or weakly activated experts, resulting in marginal performance degradation.

On the \textbf{SIPaKMeD} dataset, the trend is more distinct: performance steadily improves up to Top-6, where the model reaches its optimal accuracy (96.08\%) and F1-score (95.40\%). Beyond this point, increasing the number of experts leads to slight declines in all metrics, possibly due to unstable routing and overfitting to dataset noise.

Overall, these results demonstrate that an intermediate number of experts (Top-6) provides the best trade-off between model expressiveness and generalization, ensuring both robustness and computational efficiency across different cytology datasets.

\subsection{Effect of Axial Self-Attention}

To examine the contribution of the proposed Gaussian-enhanced axial self-attention mechanism, we conducted an ablation study by selectively removing the attention branches along the height (H) and width (W) dimensions. As summarized in \autoref{tab:axial_att}, the results clearly demonstrate that both directional attentions play complementary roles in enhancing spatial feature interactions.

On the \textbf{Mendeley LBC} dataset, removing either the horizontal or vertical attention branch leads to a noticeable performance drop, particularly in F1-score (from 99.27\% to 97.81\% or 98.15\%). When both branches are disabled, accuracy further decreases to 98.44\%, confirming that bidirectional modeling is essential for capturing the complex spatial dependencies among cytological structures. In contrast, incorporating both width- and height-wise attentions yields the best overall results, with accuracy, precision, and recall all exceeding 99.2\%.

On the \textbf{SIPaKMeD} dataset, a similar trend is observed. The complete axial attention configuration achieves the highest accuracy (96.08\%) and F1-score (95.40\%), surpassing the single-direction variants by a significant margin. The removal of either directional branch weakens the model’s ability to capture elongated or irregular spatial arrangements commonly seen in cytology images, leading to lower precision and recall.

Overall, these results validate the effectiveness of the Gaussian-enhanced axial attention design. Modeling dependencies along both axes enables the network to jointly encode global and local spatial relations, enhancing its ability to recognize subtle morphological patterns in complex cytological scenes.

\section{Conclusions}

In this study, we incorporate geometric priors and a Gaussian expert mechanism into the DFormer framework for cervical cancer screening-oriented cytology image classification. By explicitly modeling structural regularities in cervical cell images, the proposed method enhances fine-grained morphological perception and improves classification performance on public cervical cytology benchmarks. Ablation experiments validate the independent contributions and synergistic effects of the proposed modules, while comparative results demonstrate superior accuracy and robustness over existing methods. These findings suggest that geometry-aware axial attention can provide effective and interpretable decision support for automated cervical cancer screening.

\bibliographystyle{custom-unsrt}
\bibliography{sample}


\end{document}